\documentclass{article}

\PassOptionsToPackage{numbers, compress}{natbib}

\usepackage[preprint]{neurips_2022}

\usepackage[utf8]{inputenc} 
\usepackage[T1]{fontenc}    
\usepackage{hyperref}       
\usepackage{url}            
\usepackage{booktabs}       
\usepackage{amsfonts}       
\usepackage{nicefrac}       
\usepackage{microtype}      
\usepackage{xcolor}         
\usepackage[numbers]{natbib}
\usepackage{adjustbox}
\usepackage[normalem]{ulem}
\usepackage{verbatim}
\usepackage{pifont}
\usepackage{pythonhighlight}
\usepackage{multirow}
\usepackage{diagbox}
\usepackage{float}

\newcommand{\cmark}{\ding{51}}
\newcommand{\xmark}{\ding{55}}

\title{WILD-SCAV: Benchmarking FPS Gaming AI on Unity3D-based Environments}

\author{%
  Xi Chen \\
    Inspir.AI \\
    \texttt{chenxi@inspirai.com}
   \And
   Tianyu Shi \\
   University of Toronto \\
  \texttt{ty.shi@mail.utoronto.ca} \\
   \AND
   Qingpeng Zhao \\
   Nanjing University \\
  \texttt{mf20150077@smail.nju.edu.cn} \\
   \And
   Yuchen Sun \\
   Inspir.AI \\
  \texttt{
sunyuchen@inspirai.com} \\
   \And
   Yunfei Gao \\
   Inspir.AI \\
  \texttt{
claude@inspirai.com} \\
   \And
   Xiangjun Wang \\
   Inspir.AI \\
   \texttt{
xj@inspirai.com} \\
}

\begin{document}

\maketitle

\begin{abstract}
Recent advances in deep reinforcement learning (RL) have demonstrated complex decision-making capabilities in simulation environments such as Arcade Learning Environment~\cite{bellemare2013arcade}, MuJoCo~\cite{todorov2012mujoco} and ViZDoom~\cite{kempka2016vizdoom}. However, they are hardly extensible to more complicated problems, mainly due to the lack of complexity and variations in the environments they are trained and tested on. Furthermore, they are not extensible to an open world environment to facilitate long-term exploration research. To learn realistic task-solving capabilities, we need to develop an environment with greater diversity and complexity. 
We developed WILD-SCAV, a powerful and extensible environment based on a 3D open-world FPS (First-Person Shooter) game to bridge the gap. It provides realistic 3D environments of variable complexity, various tasks, and multiple modes of interaction, where agents can learn to perceive 3D environments, navigate and plan, compete and cooperate in a human-like manner. WILD-SCAV also supports different complexities, such as configurable maps with different terrains, building structures and distributions, and multi-agent settings with cooperative and competitive tasks. The experimental results on configurable complexity, multi-tasking, and multi-agent scenarios demonstrate the effectiveness of WILD-SCAV in benchmarking various RL algorithms, as well as it is potential to give rise to intelligent agents with generalized task-solving abilities. The link to our open-sourced code can be found here\footnote{\url{https://github.com/inspirai/wilderness-scavenger}}.

\end{abstract}

\section{Introduction}

A challenging benchmark is essential to fully assess the capabilities of deep reinforcement learning (RL) algorithms. Many simulators have been used to challenge and evaluate RL algorithms, bringing significant contributions and development to the RL community. For example, the Arcade Learning Environment (ALE) \cite{bellemare2013arcade} presents a valuable test-bed with a collection of Atari 2600 games to evaluate RL algorithms. Deep Q Network (DQN) successfully learns to play the Atari 2600 games given screen pixels as inputs and achieves outstanding performance on ALE ~\cite{mnih2013playing}. Other examples, e.g., Dueling DQN~\cite{wang2016dueling}, Prioritized DQN~\cite{schaul2015prioritized}, Quantile Regression DQN~\cite{dabney2018distributional}, also largely benefited from ALE as they were developed. To support more general task scenarios, Gym~\cite{brockman2016openai}, built by OpenAI, exposes standard interfaces for testing different RL algorithms over a variety of environments, allowing researchers to easily compare the performance of alternative approaches.

Recently, RL methods have demonstrated superior ability to humans on multiple Atari tasks. DeepMind built Agent57~\cite{badia2020agent57}, which is the first deep reinforcement learning model that surpasses human baseline performance on all Atari games. The success of Agent57 comes from its meta-controller design to balance efficient exploration and exploitation. As we are building more and more intelligent task-solving agents, it appears that memory utilization~\cite{kapturowski2018recurrent}, curiosity seeking~\cite{badia2020never}, and exploration-exploitation balancing~\cite{badia2020agent57} have become some of the key points to motivate these developments.

To explore the potential of RL algorithms and push forward the performance limits, the community keeps seeking more challenging environments that generally involve real-world tasks and corresponding interactions. For example,  navigation and resource collection tasks ~\cite{alonso2020deep} are very common in the real world and are integrated into several environments, e.g.,  Coin-Run~\cite{cobbe2019quantifying}, Deepmind Lab~\cite{beattie2016deepmind}. Furthermore, by introducing more comprehensive forms of observational inputs and more flexible action types, agents can be evaluated on not only navigation and resource collection but also more complex tasks like the move-and-shoot task in ViZDoom~\cite{kempka2016vizdoom}. 
While adding action options may lead to an exponential increase in the complexity of the policy, several advanced RL algorithms have been proposed to solve these challenges. For instance, hierarchical Reinforcement Learning (HRL)~\cite{song2019playing} has been proposed to solve the huge action space and exploration challenges, in which the high-level layer learns policy through options while the low-level layer executes basic options, such as motion, attack, tool, and resources.


Following the success of ViZDoom, training intelligent agents with general task-solving capabilities in open-world environments has attracted increasing attention. However, the lack of adequate training and testing benchmark environments remains an obstacle to research in this field. ViZDoom is a useful and fast 3D environment but still oversimplified in both the visual style and the world composition compared to the simulation appearance in modern 3D games, which is relatively detached from the more practical side of the community's needs. 
In this context, the goal of WILD-SCAV is to advance research in the field of open-world intelligent agent learning. As a stepping stone to the ultimate goal of learning highly intelligent agents' ``living'' in the virtual world, we decided to first focus on open-world FPS games. With the popularity of battle royale games, such an environment can serve as a flexible and open playground for both algorithmic research in RL and solution optimization in the game industry.

To the best of our knowledge, WILD-SCAV is the first FPS-based environment that allows agents to explore in such a 3D world environment. Our main contributions are listed as follows:

\begin{itemize}
    \item WILD-SCAV  provides a customizable 3D environment, where the ground landscape, the structure of houses, and the placement of various types of objects can be generated with PCG (Procedural Content Generation) technique~\cite{togelius2013procedural}, which is very suitable for NPC research~\cite{alonso2020deep}.
    
    \item WILD-SCAV focuses on learning intelligent agents in the open-world environment. It provides more spaces for exploration, more flexibility in task design and open challenges for generalization with diverse learning scenarios.
    
    \item WILD-SCAV supports agent training in multi-task and multi-agent scenarios, similar to the common setup in recently popular battle royale games (e.g., PUBG~\cite{wiki:PUBG:_Battlegrounds}), which includes tasks like random target navigation, competitive/cooperative resources gathering, and free fight etc. 
    
    

\end{itemize}








\section{Background: 3D open-world FPS game environment}

AI-powered next-generation gaming experiences for open-world games have attracted increasing attention after the success of AI in StarCraft II~\cite{vinyals2017starcraft} and DOTA~\cite{xia2018dota} as the next big challenge. However, the lack of satisfactory testing environments remains an obstacle to research in this area. In this context, the goal of this competition is to promote research on intelligent agent learning for 3D open-world FPS games. In 3D open-world environments, an agent perceives its environment in a human-like manner, using visual scenes as input. Intelligent agents are expected to integrate visual perception and contextual game features, process incomplete information, deal with the dynamic variation of environments and multiplayer enemies, and then perform long-term planning. In addition, to maximize scores on the tasks set, agents must generalize their learned skills to unknown test environments. For the competition, we provide an FPS game environment similar to popular Battle Royale games (e.g., PUBG), where multiple players compete against each other for limited supply resources. We will evaluate each trained agent on the randomly generated battlegrounds.


\section{WILD-SCAV Environment}

\subsection{Observation Space}

The gameplay interface provides multiple sources of information about the agent (e.g., location, orientation) and its surrounding environment. The observation mainly consists of two parts, visual perception inputs and game variables.

\begin{figure*}[ht!]
    \includegraphics[width=.5\textwidth,height=.3\textwidth]{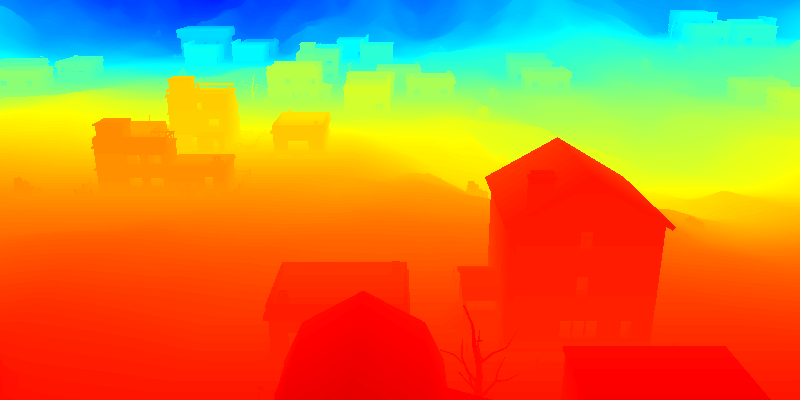}\hfill
    \includegraphics[width=.45\textwidth,height=.3\textwidth]{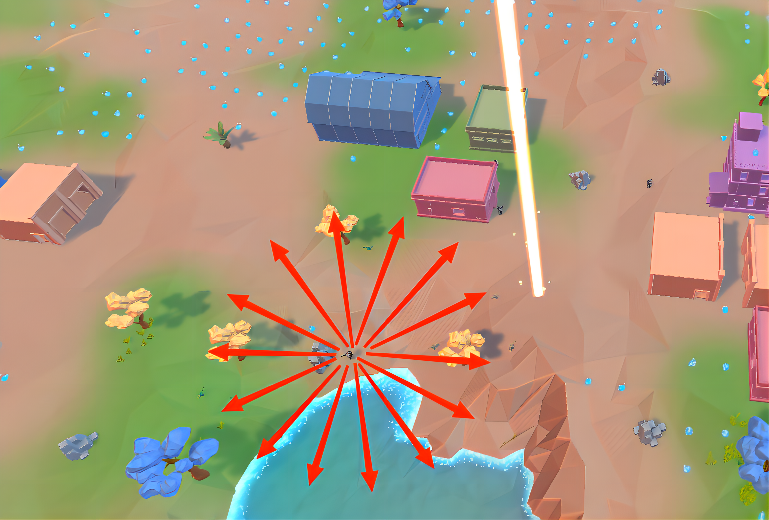}
    \caption{\textbf{left:} depth map visualization example. \quad \textbf{right:} schematic diagram of LIDAR perception. The arrows means the depth map is a 360-degree panorama.}
    \label{perceptron}
\end{figure*}

\begin{figure*}[ht!]
    \centering
    \includegraphics[width=0.23\textwidth]{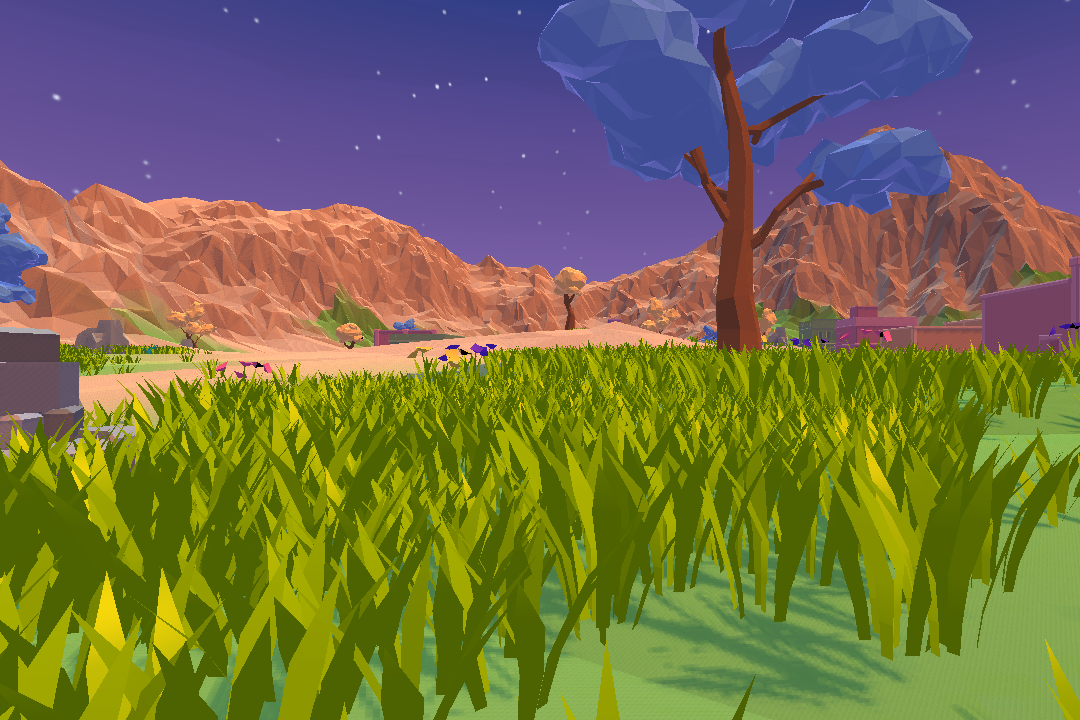}
    \hfill
    \includegraphics[width=0.23\textwidth]{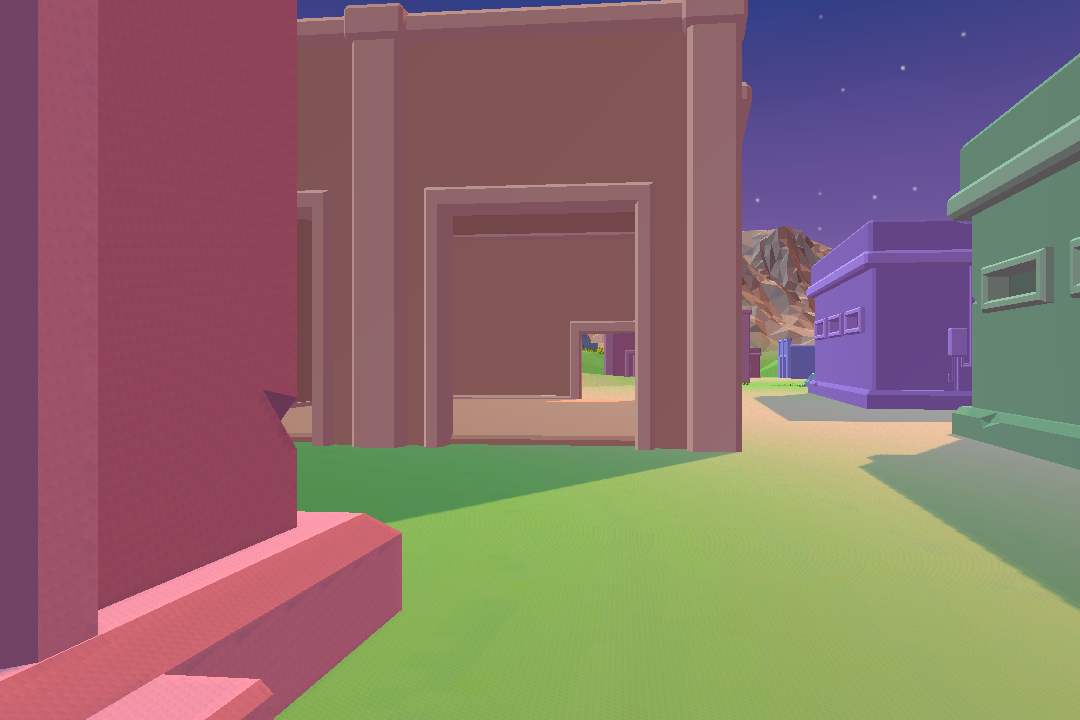}
    \hfill
    \includegraphics[width=0.23\textwidth]{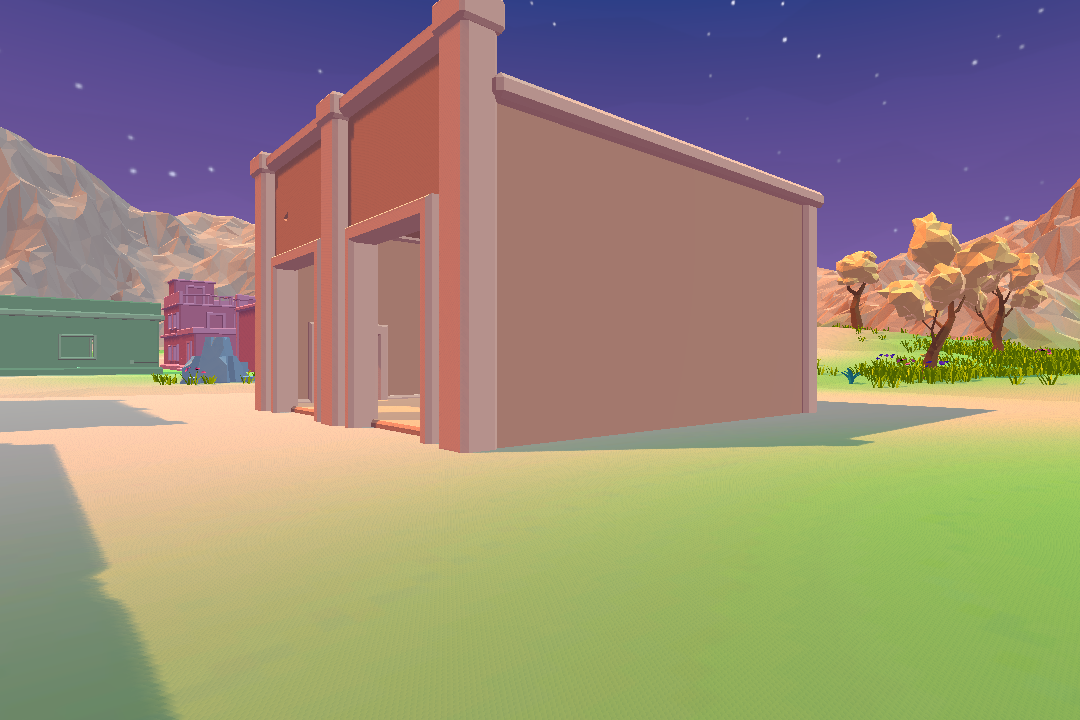}
    \hfill
    \includegraphics[width=0.23\textwidth]{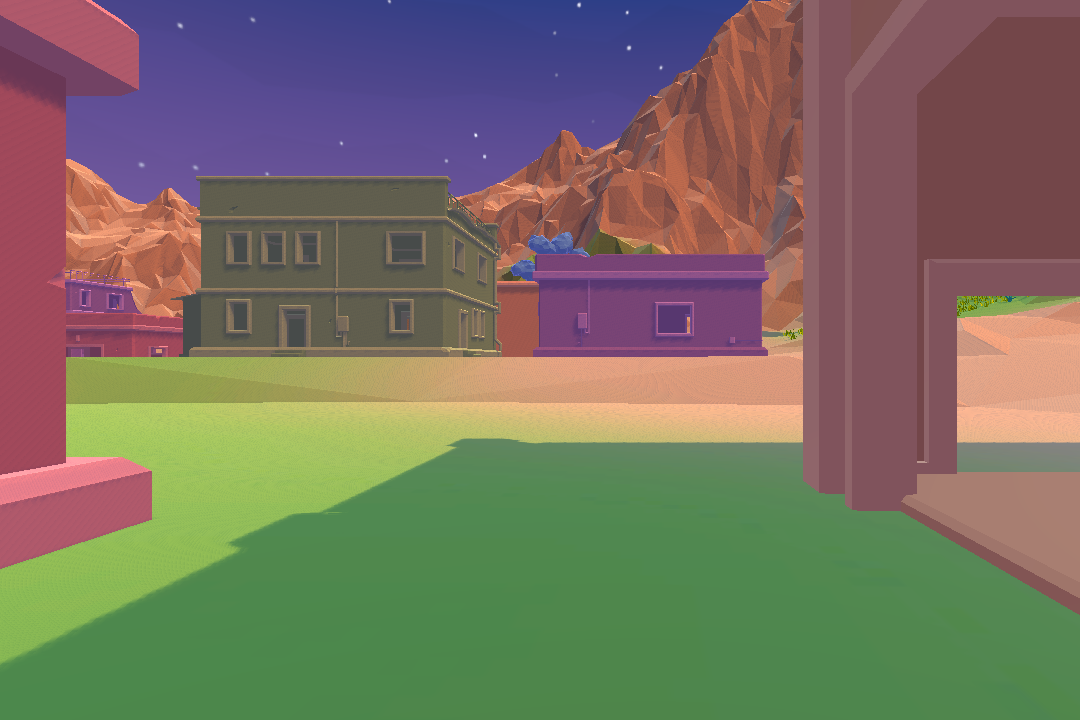}
    \caption{Images of the agent's first person camera from different viewing angles.}
    \label{camera}
\end{figure*}

\textbf{Visual perception.}
We implemented an efficient way to compute a low-resolution depth map (e.g. $10 \times 10$) from the agent's camera using only the location, orientation values, and mesh data of the static scenes, to enable efficient feature learning in the scenario of large 3D open worlds. In this way, we trade-off between granularity of visual information and computation efficiency to allow researchers to 
get most out of this new environment when using it for agent training. However, to make our environment more usable for more generalized tasks, we also added support for calculating high-resolution depth maps, LIDAR detection, and the first person camera images, as shown in Figure~\ref{perceptron} and Figure~\ref{camera}. A short video visualization of the camera images can be found here.\footnote{\url{https://youtu.be/VrXPF8HoIbU}}

\textbf{Game variables.} We also provide access to multiple game variable information. The variables include current location and orientation of the agent, state of motion (\emph{e.g.} \emph{on ground} or \emph{in the air}), health, state of combat (\emph{e.g.} \emph{firing}, \emph{being hit}), and task-related metrics (\emph{e.g.} target location, number of collected resources). These game variables can be essential to task-solving in more complex tasks. 

\vspace{3mm}

\vspace{3mm}

The above observations provide rich and efficient information for the agent to learn optimal policies. Users have the freedom to construct customized observational features for different training tasks.


\subsection{Action Space}
WILD-SCAV provides comprehensive and structured action spaces. Users can easily explore the action spaces to facilitate skill learning of the agent. To be more specific, it support actions for \emph{navigation} task, such as walking direction (\texttt{WALK\_DIR}); walking speed (\texttt{WALK\_SPEED}). It also supports camera angle direction change for wider range of observation, such as horizontal camera angle (yaw) between two frames (\texttt{TURN\_LR\_DELTA}) and vertical camera angle (pitch) between two frames (\texttt{LOOK\_UD\_DELTA}). 

Unlike previous shooting game environments ~\cite{kempka2016vizdoom}, our environment has more complex terrain, where agents have huge space of movement trajectories when moving around. An agent can jump up stairs (\texttt{JUMP}) and pick up randomly spawned resources (\texttt{PICKUP}). These characteristics of WILD-SCAV make it a favorable testbed for cooperative and competitive resource gathering tasks.

In addition, in contrast to multi-agent cooperative and competitive environments like Multi-Agent Particle environment~\cite{lowe2017multi}, we also introduce the combat game system to improve the complexity of the environment. There are two combat actions: (a) whether to fire the weapon and cost one bullet at the current time step (\texttt{SHOOT}) and (b) whether to refill the weapon's clip using spare ammo (\texttt{RELOAD}). The detailed tasks illustration will be analyzed in \ref{tasksillustra} and a summarized action space description for each task is shown in Table \ref{tab:action_space}.

    






    
    
    
    




\begin{table}[htbp]
    \centering
    \label{tab:action_space}
    \begin{tabular}{cccccc}
    \toprule
    Action Class & Navigation & Supply Gathering & Supply Battle & Type & Range\quad \\
    \midrule
    WALK\_DIR                 & \cmark         & \cmark             & \cmark  & float & $[0., 360.]$ \\
    WALK\_SPEED                   & \cmark         & \cmark             & \cmark  & integer & $[0, 10]$\\
    TURN\_LR\_DELTA               & \cmark         & \cmark             & \cmark  & float & $[-\infty, \infty]$\\
    LOOK\_UD\_DELTA               & \cmark         & \cmark             & \cmark  & float & $[-\infty, \infty]$\\
    JUMP                        & \cmark         & \cmark             & \cmark   & boolean & True / False\\
    PICKUP                      & \xmark         & \cmark             & \cmark  & boolean & True / False\\
    SHOOT                       & \xmark         & \xmark             & \cmark & boolean & True / False\\
    RELOAD                    & \xmark         & \xmark             & \cmark  & boolean & True / False\\
    \bottomrule
    \end{tabular}
    \vspace{2mm}
    \caption{Action space description for the typical tasks }
\end{table}



\subsection{Typical Tasks}
\label{tasksillustra}

In this section, we will discuss some typical tasks supported by the WILD-SCAV environment. The basic tasks are navigation, supply gathering, and battle. By combinations of different typical tasks, different experiment scenarios can be created to validate the performance of different RL algorithms.

\subsubsection{Navigation}
The challenge for the agent is to navigate as quickly as possible from a starting location to a destination (both randomly selected) in a randomly generated open world. The world consists of various structures, such as buildings, trees, rocks, and lakes. Ideally, an efficient and general navigation strategy must be able to skillfully use these elements and flexibly adapt to new worlds that have not been seen yet.

\begin{itemize}
    \item \textbf{State:} The observation is the target position, current position and the depth image.
    \item \textbf{Action:}  Walking direction, walking speed, camera angle (yaw, pitch), and jump.  
\item \textbf{Reward:} The agent is rewarded when it reaches the target point.

\item \textbf{Evaluation criteria:} Time consumed to reach the target, i.e., the episode length.

\end{itemize}

\begin{figure}[htbp]
    \centering
    \includegraphics[width=0.9\textwidth]{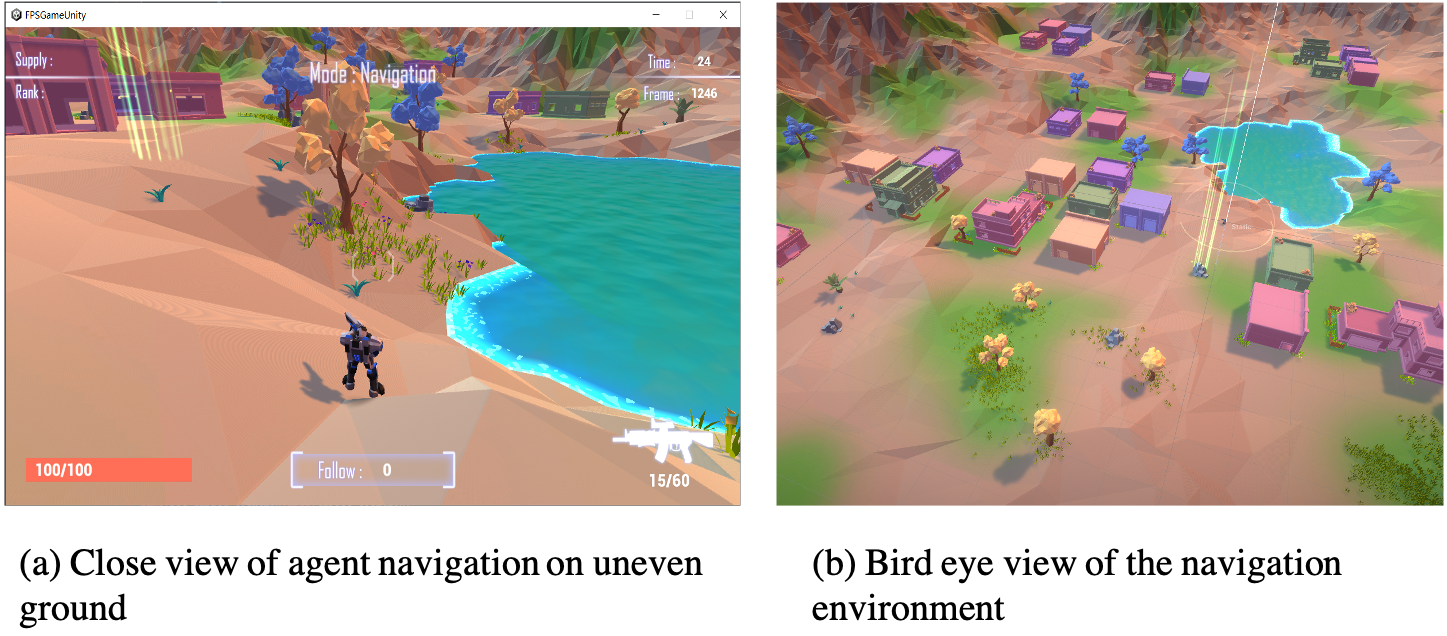}
    \caption{Navigation task: The agent is trying to navigate to the target location}
    \label{navi}
\end{figure}

\subsubsection{Supply Gathering }
The challenge for the agent is to collect as many supplies as possible by opening the blue supply boxes in a randomly generated open world with unknown supply distribution. Supply boxes may appear at any accessible location in the open world, for example, on outdoor grounds or on a certain floor of a house, hiding behind a tree or a stone. To make it even more challenging, we have designed a special mechanism to determine the number of supplies stored in each supply box. In general, a supply box inside a building contains a significantly higher number of supplies than an outdoor box, while the number of supply boxes inside buildings could be relatively small. These agents are encouraged to explore both outdoor areas and indoor spaces to maximize their collection of supplies. Also, the actual supply quantity in each box has some randomness but is constrained within given ranges.

\begin{figure}[htbp]
    \centering
    \includegraphics[width=0.9\textwidth]{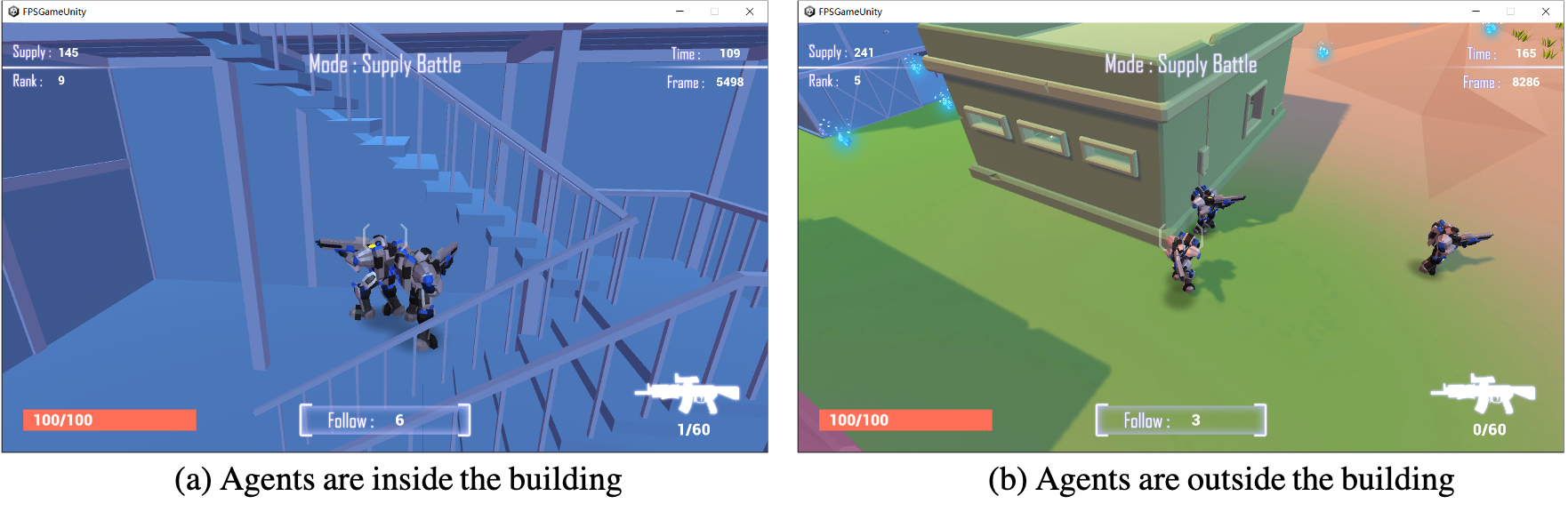}
    \caption{Supply gathering task 1: two groups of agents searching supplies in the housing area}
    \label{coop}
\end{figure}

\begin{figure}[htbp]
    \centering
    \includegraphics[width=0.9\textwidth]{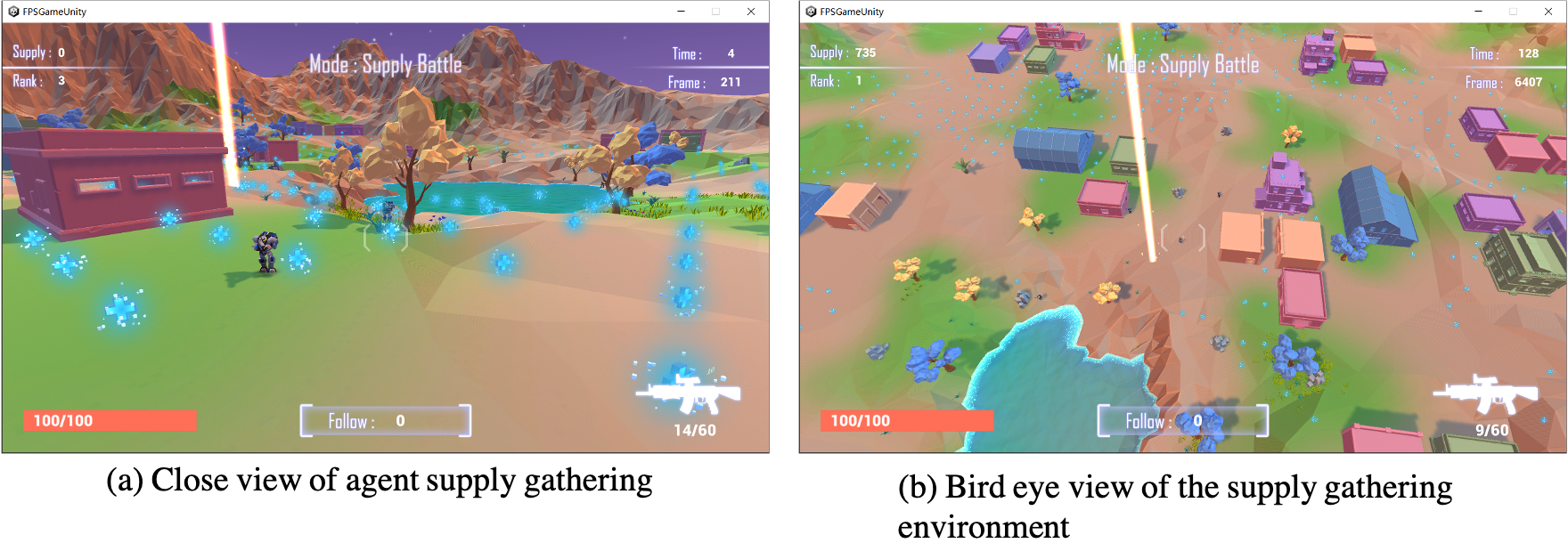}
    \caption{Supply gathering task 2: the agent team trying to collect supplies as fast as possible}
    \label{coop}
\end{figure}

\begin{itemize}
    \item \textbf{State:} The observation are the locations of all nearby supplies, the specific number of supplies at each location, the agent's current location and the depth image.
    \item \textbf{Action:}  Walking direction, walking speed, camera angle (yaw, pitch), and jump.  
\item \textbf{Reward:} To add variability to the task, we can set different termination criteria. We design two separate sub tasks. In the first task, the agent will get a high reward if it can collect supply as many as possible. In the second task, the agent is rewarded for collecting the target supply as fast as possible. 

\item \textbf{Evaluation criteria:} In the first sub-task, we use the number of collected supplies to evaluate. In the second sub-task, we use episode length as evaluation criteria.

\end{itemize}

\subsubsection{Supply Battle}
Based on the supply gathering, multiple agents are dropped into a large, randomly generated world (terrain, buildings, plants, supplies) and compete for supplies. These agents can fight with weapons and respawn when killed. During combat, agents are encouraged to take cover to ensure their safety. When an agent dies, some of his supplies drop and can be collected by other agents. Since the total number of supplies in the game is limited, agents should learn to optimize their search strategy to collect supplies faster. With the introduction of the combat system, agents can also use their combat advantage to snatch supplies from other agents to speed up the accumulation of supplies.

The observed state and actions of the agent are very similar to the material collection task of the previous task or are added based on it. In addition to its own attributes in the gathering task, observation can also include the state of other agents in the field of view, but if the agent is not in the field of view, this state cannot be obtained, which leads to the non-stationarity and partial observation of the environment, making it more complex and difficult to learn. In addition to basic actions such as movement, we further extend the action types by adding the behavior of shooting attacks and changing bullets. It is expected to learn a strategy that can obtain their supplies by killing other agents and also avoid being killed by other agents.

\section{Experiments}
\label{exp}
In this section, we present the results and analysis of the experiments based on the combination of the elements of the typical task from the above section~\ref{tasksillustra}. Note that we could extend more combinations of different typical tasks, but for this paper, we decided to focus on several representative experiments and demonstrate how our environment can be configured to assess the performance of different RL methods across different learning scenarios.

For a fair comparison, we adopt the same resource configuration and fixed hyper-parameters for all tasks and experiments to test the adaptability and scalability of the compared algorithms. We use the same machine with 96 Intel(R) 8163 (2.50GHz) CPU cores to run all algorithms. For simplicity and better reproducibility, we construct all training pipelines using the standard trainer APIs from the commonly used RLlib~\cite{liang2018rllib} among the research community.

To keep it simple and clear, for the benchmark experiments in this paper, we only select three commonly used RL algorithms, including IMPALA~\cite{espeholt2018impala}
Proximal Policy Optimization (PPO)~\cite{schulman2017proximal} and Asynchronous Advantage Actor-Critic (A3C)~\cite{babaeizadeh2016reinforcement}. For each episode rollout during training, while it is possible to let the agent explore the environment for a long time, we can definitely speed up the learning by setting appropriate episode lengths $T$ for different tasks. In our experiments, we empirically choose the episode length for each task. Specifically, we fix the maximum episode length to $T=400$ for the single-agent navigation and the multi-agent target capture task, and we fix the episode length to $T=600$ for the multi-agent supply gathering task.


\subsection{Single-Agent Navigation}
For an open-world 3D environment, the navigation task is the basis for completing other more complex tasks. For navigation tasks, even if the environment is very complex, the agent can quickly learn a relatively good strategy to reach any target point thanks to the easy randomization of the start location. The agent is given a reward of $+1$ if it reaches the target, and $0$ otherwise. However, such a reward distribution can be very sparse, especially for large map sizes.



\begin{figure}
    \centering
    \includegraphics[width=1\linewidth]{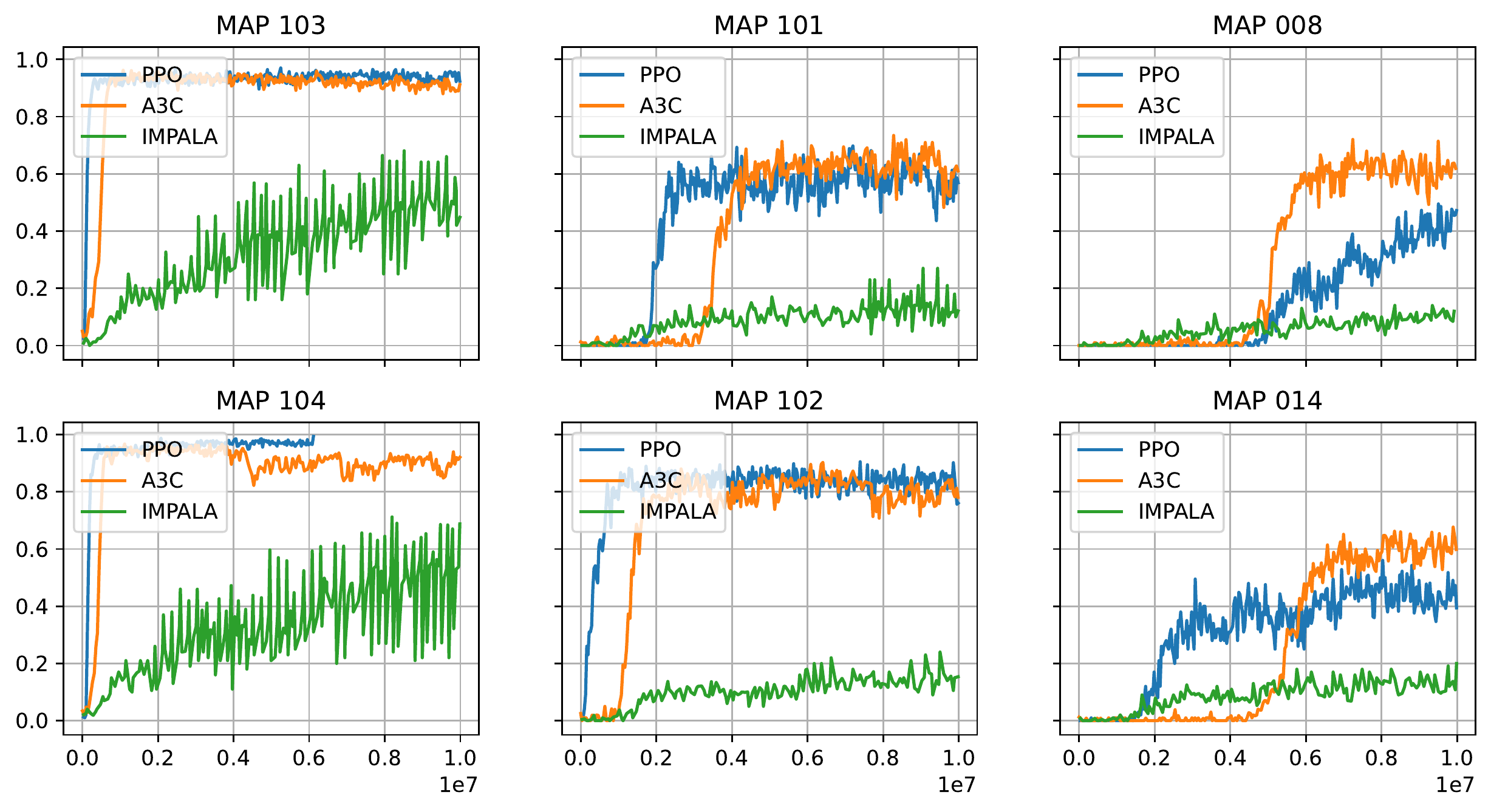}
    \caption{Average episode rewards during training on different sizes of maps. Small ($100\times 100$): 103, 104; Medium ($200\times 200$): 101, 102; Large ($500\times 500$): 008, 014.}
    \label{learningcurve}
\end{figure}

Figure \ref{learningcurve} shows the task training curve of the evaluated algorithms for different map sizes and scene complexity. Overall, for tasks of different complexity and different sizes, all algorithms can quickly find some decent strategies to reach the target point, and they will always slowly explore new strategies to reach the target in faster ways.


Seen from the results in Figure~\ref{learningcurve} and Table~\ref{navigation}, with large size and higher scene complexity of the map, the average episode reward becomes smaller, resulting in more ``slow-learning'' steps. To help improve the learning efficiency, one promising approach is to design better Intrinsic curiosity-driven reward to encourage more efficient exploration~\cite{hare2019dealing}. 

In addition, both A3C and PPO can learn a good strategy, while the performance of IMPALA is relatively poor. PPO performs best in maps with a small size and shows the highest improvement speed at the early stage, while A3C performs better under larger sizes of the map and may need longer time steps to reach a high-performance level. IMPALA in this task indeed shows steady growth of performance but has a much slower speed and higher fluctuation.

\begin{table*}[ht]
\label{navigationtable}
\centering



\begin{tabular}{c|lc|lc|lc}

\toprule
\multicolumn{3}{c}{\textbf{A3C}} & \multicolumn{2}{c}{\textbf{PPO}} & \multicolumn{2}{c}{\textbf{IMPALA}} \\
\midrule
 Map & 	EpsLen & SuccRate  & EpsLen & SuccRate & EpsLen & SuccRate  \\ %
\midrule
101  & 219.4 (132.8)  & 0.73  & 228.4 (140.1)   & 0.69 & 330.9 (124.4 ) & 0.27\\
102  & 185.2 (111.5)  & 0.90  & 170.6 (115.0)   & 0.90 & 330.5 (119.8)  &  0.29\\
103  & 158.2  (85.4)   & 0.96  & 87.5 (73.6)  & 0.97 & 136.8  (130.8)  & 0.82\\
104  & 118.8  (77.9)  & 0.96  & 63.9 (45.4)   &0.99 & 109.5  (122.5)  & 0.86\\
008    & 255.9  (126.9) & 0.71  & 283.1  (118.6) & 0.66 & 354.1  (105.0) & 0.19\\
014   & 239.6 (124.3)  & 0.74 & 284.0  (121.7) &0.62 & 353.4 (97.8)  & 0.23\\
\bottomrule

\end{tabular}

\caption{Evaluation results on the single agent navigation task. We compare average episode lengths and success rates achieved with the learned policies on different maps. For the episode length, we report both the mean value and the standard deviation (in the brackets).} \label{navigation}

\end{table*}



\subsection{Cooperative Supply Gathering.}
The second task is a \emph{multiplayer cooperative} task. In this task, multiple agents are taking action as a team to collect as many supplies as possible within an episode. In this task, the agent is trying to learn a policy that jointly considers their own states (e.g., current agent location), the supply state (e.g., location of a nearby supply), and the teammate state (e.g., location of other agents) to collect the supplies in a more efficient way than the single-agent scenario. We run all the experiments on map 101, where randomly generated supplies are distributed denser near the center of the map and sparser far away from the center. For each supply collected, the agent is given a small positive reward.

Figure~\ref{coop1} shows the change in the average number of supplies collected in an episode during training. We can see that PPO learns to handle the task quite fast, with only $2e6$ time steps when reaching an acceptable level of performance. IMPALA has the lowest improvement speed at the start of training but shows a relatively steady and moderate speed of performance growth. However, A3C performs not very well and seems to require more time steps of experience to basically handle the task.
The results may have shown the low exploration efficiency of A3C in case of sparse reward distribution and that PPO can be a favorable choice in such an open-world environment.

In such a multi-agent case, how to coordinate the actions of all agents in order to collect the most supplies in a limited time is still a challenging problem. Although here we only test the performance of three commonly used algorithms, some recently proposed MARL algorithms~\cite{mao2022improving,yang2020qatten} can be simply applied with little modification of the environment. From the results in Table~\ref{coop11table}, we can see that the best method, PPO, can only reach a level of around 100 average supplies while there are more than 200 supplies on the 
entire map. This may be due to the exploration difficulty that agents are more likely to collect supplies in a small sub-region where supplies are relatively close to the agents' current locations other rather than explore large open areas. But some ``treasure'' places with more densely distributed supplies can be far away from the start locations. As the number of agents increases, supplies in a sub-region can be quickly depleted by a small group of agents, leaving other agents a sparser reward distribution, which suggests that the environmental setups should be more subtle on the multi-agent task when the number of agents increase, to avoid insufficient local optimal.


    
\begin{table*}[htbp]
    \centering

    \begin{tabular}{c|c|c|c}
    \toprule  
    \multicolumn{2}{c}{\textbf{A3C}} & \multicolumn{1}{c}{\textbf{PPO}} & \multicolumn{1}{c}{\textbf{IMPALA}} \\
    
    \midrule
    Map & SupplyNum    &   SupplyNum     & SupplyNum \\
	\midrule
    101 & 49.49 (18.67)  &   98.92 (52.62) & 91.68 (45.90) \\
    \bottomrule
    
    \end{tabular}
    
    \caption{Evaluation results on the cooperation supply gathering task. We report both the mean and standard deviation (in the brackets) of the total collected supply numbers in 3 minutes with 4 agents.}\label{coop11table}
    
\end{table*}


    
\subsection{Cooperative Target Capture}    
The last task is also a \emph{multiplayer cooperative} task, where multiple agents are trying to capture the target supply as fast as possible cooperatively. The supply is hidden at some point in the open world. Once an agent captures it, the episode is considered a success and finished, and a $+1$ reward is given to this agent.

\begin{figure}[htbp]
\centering
\begin{minipage}[t]{0.48\textwidth}
\centering
\includegraphics[width=\linewidth]{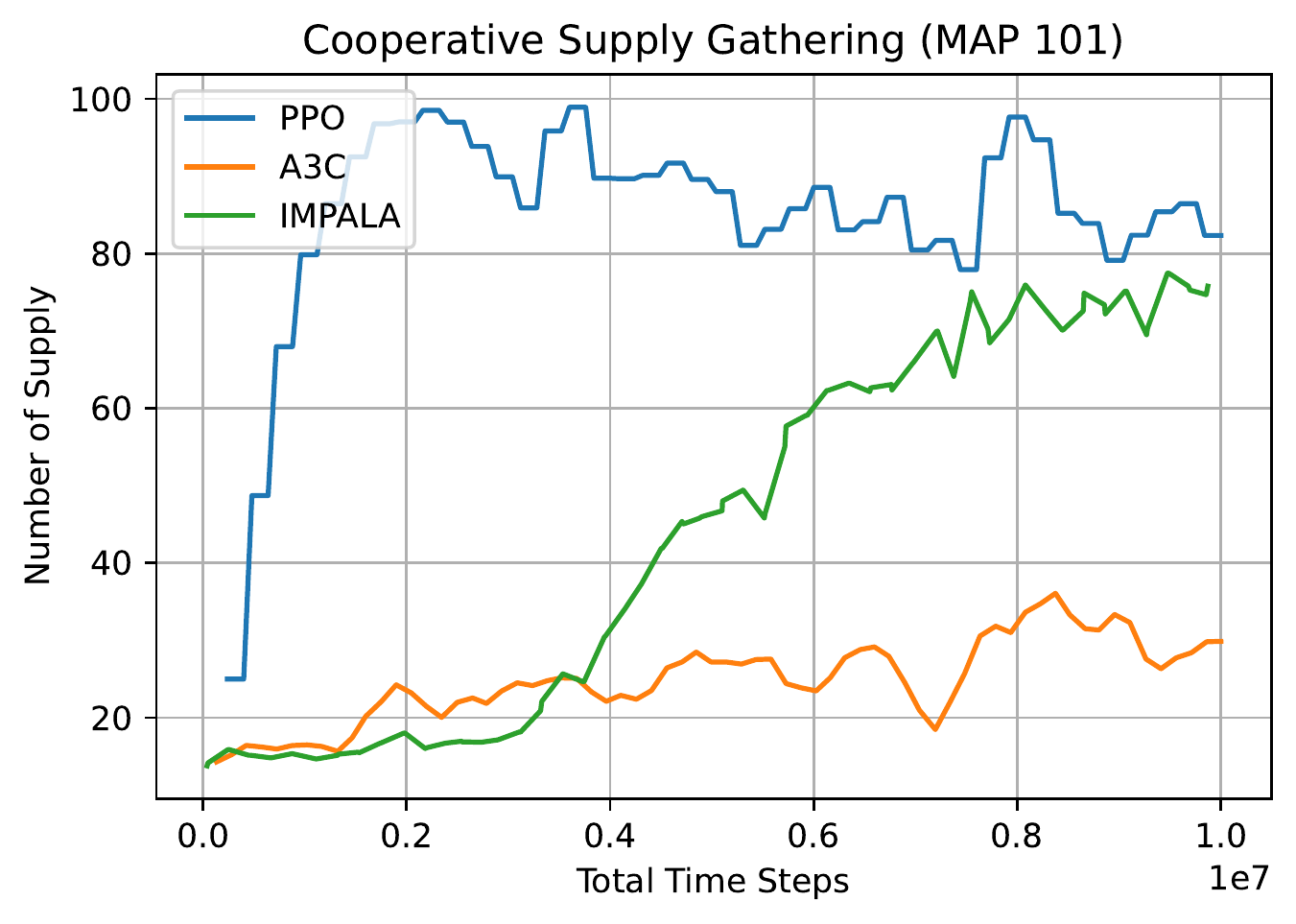}
\caption{Average episode rewards during training on the supply gathering task. Five agents are randomly generated in the map 101. Each agent receives $+1$ reward for each supply collected within the episode time limit. The agent team tries to cooperatively collect supplies as many as possible.}
\label{coop1}
\end{minipage} 
\hfill
\begin{minipage}[t]{0.48\textwidth}
\centering
\includegraphics[width=\linewidth]{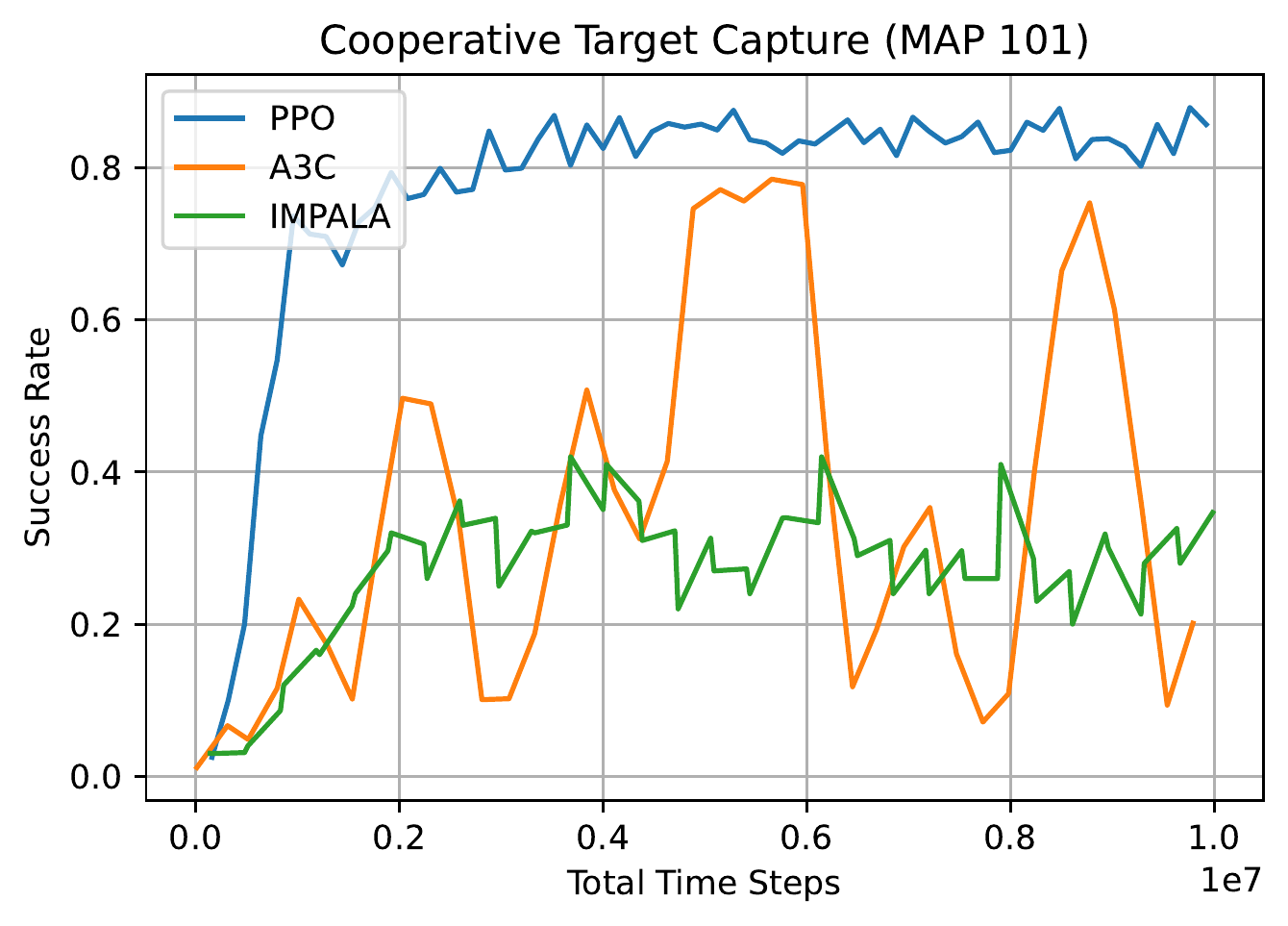}
\caption{Average episode reward during training on the cooperative target capture. The agent team is encouraged to capture the target supply as fast as possible}
\label{coop2}
\end{minipage}
\end{figure}

In Figure~\ref{coop2}, we show the learning results of different methods where PPO can achieve the highest reward. When collecting the state and supply position of each agent, the strategy will assign the nearest agent to quickly catch this supply. For the same map configuration, such a multi-agent environment can solve the task faster than a single-agent environment compared to the results in Table~\ref{navigationtable}. Similar to previous experiments, we report the mean and standard deviation in the bracket.
Note that the standard deviation is a bit high, which is because we randomly generate the agent in the map without fixed agents' start and target locations. To conclude, we find that PPO once again achieves the best performance with the smallest episode length.

\begin{table*}[htbp]
    \centering
    
    \begin{tabular}{c|c|c|c}
    
    \toprule
    \multicolumn{2}{c}{\textbf{A3C}} & \multicolumn{1}{c}{\textbf{PPO}} & \multicolumn{1}{c}{\textbf{IMPALA}} \\
    
    \midrule
    Map  & EpsLen           &   EpsLen       & EpsLen \\
    \midrule
    101  & 143.18 (122.71)   & 94.69 (98.55) & 251.58 (166.72) \\
	\bottomrule
    
    \end{tabular}
    
    \caption{Evaluation results on the multi-agent target capture task. We report both the mean value and standard deviation of the episode length to reach the target location.}\label{coop2table}
    
    \end{table*}

\section{Related work}

In general, modern 3D FPS games are inherently incomplete information games that are extremely hard to learn winning strategies in multiple-player scenarios and are known to have no optimal policy. Despite the difficulties, there have been attempts over the past decade to apply reinforcement learning in FPS games. To our best knowledge, the most influential work is the Augmented DRQN model~\cite{lample2017playing}, where the method leverages both visual input and the game feature information (e.g., presence of enemies or items) and modularizes the model architecture to incorporate independent networks to handle different phases of the game. Their approach successfully learned a competitive FPS agent by minimizing a Q-learning objective and showed better performance than average human players. Following this success, more work on learning FPS game agents has been proposed, such as Arnold~\cite{chaplot2017arnold} which benefits from the Action-Navigation architecture, Divide and Conquer deep reinforcement learning~\cite{papoudakis2018deep} which further refined the idea of separating the control strategies of map exploration from enemy combat. Although these methods have shown promising results, their training and evaluation context is largely limited to old-fashioned video games with relatively small world sizes, such as VizDoom (originally 1993) and Quake 3 (originally 1999). Recently, Pearce and Zhu ~\cite{pearce2021counter} tried to learn an FPS agent to play CSGO, a phenomenal modern 3D FPS game with high-resolution visual rendering. This new game environment not only introduces more computational burden (mostly due to extracting visual features) but also makes it more difficult for the agent to explore and adapt to the game world efficiently. The new approach addressed the challenge primarily by using behavioral cloning, and the learned agents showed reasonably good performance compared to normal human players in the Deathmatch mode. Note that there are also other 3D simulators, such as Mujoco~\cite{todorov2012mujoco}, DeepMind Lab~\cite{beattie2016deepmind}, etc. They are not extensible to more complicated real-world problems. On the other hand, our simulator is more suitable for open-world exploration. 
Recently, MINEDOJO has been developed with thousands of diverse open-ended tasks~\cite{fan2022minedojo}. With MINEDOJO's data, one can leverage large pre-trained video language models to learn reward functions and then guide agent learning in various tasks. Building upon these works, we seek to further expand the frontiers of intelligent agent learning in modern large-scale open-world games.

\section{Future work }

As discussed in Section~\ref{tasksillustra}, we can create great combinations of different tasks (i.e., navigation, supply gathering, battle) to support various experiments. Besides these representative experiments, as discussed in  Section~\ref{exp}, we also tried other experiments. However, we find that traditional methods, as discussed in Section~\ref{exp}, will fail in more challenging experiments. For example, we randomly generated ten agents on map 103, as shown in Figure~\ref{mapviz}. These agents are controlled by the random policy while we train another agent, which is controlled by A3C. However,  we find that the trained agent is hard to learn an appropriate behavior because it cannot go to the target point successfully.

We suggest that there are a few potential research ideas for future improvement. Firstly, when the map is becoming larger, the agent will become hard to find an enemy, which will be a barrier to efficient exploration. Curiosity-driven reinforcement learning method can be utilized to encourage exploration in sparse reward setting~\cite{still2012information,zheng2021episodic}. Secondly, we find that when the agents' number increase, the learned agent is hard to process in such a large observation space, and the learning becomes very difficult. In the future, we can investigate how to better capture the mutual interplay of different agents through communication, such as graph neural network~\cite{shi2020efficient}. Lastly, as  WILD-SCAV can support Procedural Level Generation (PCG), it is also worth studying PCG-based reinforcement learning to improve the generalizability of our agent~\cite{justesen2018illuminating}.

\section{Conclusion}

In this paper, to bridge the gap with realistic NPC gaming problems, we present WILD-SCAV, the first realistic 3D FPS-based environment, with the support for configurable complexities, multi-tasks, and multi-agents. Built upon PCG world generation techniques, WILD-SCAV enables the configurations of object types, areas, locations, orientations, etc. We have created several demo tasks, i.e., navigation, supply gathering, and supply battle, to evaluate the performance of different RL methods and enable researchers to develop more powerful algorithms through configurable environments. To facilitate further research development, we open-sourced our code for the simulator and training agents with A3C, PPO, and IMPALA. We also host the open-world FPS game AI competition to attract global researchers to innovate on the algorithms. Winners will be selected based on the evaluation of randomly generated environments. We believe WILD-SCAV could further push forward the development of AI algorithms in the 3D Open World, bringing it a step closer to intelligent and generalized task-solving agents in realistic NPC research.
\bibliographystyle{unsrt}
\bibliography{refs}

\appendix

\section{Appendix}

\subsection{Generating benchmark maps with PCG}
\begin{figure}[htbp]
    \centering
    \includegraphics[width=120mm]{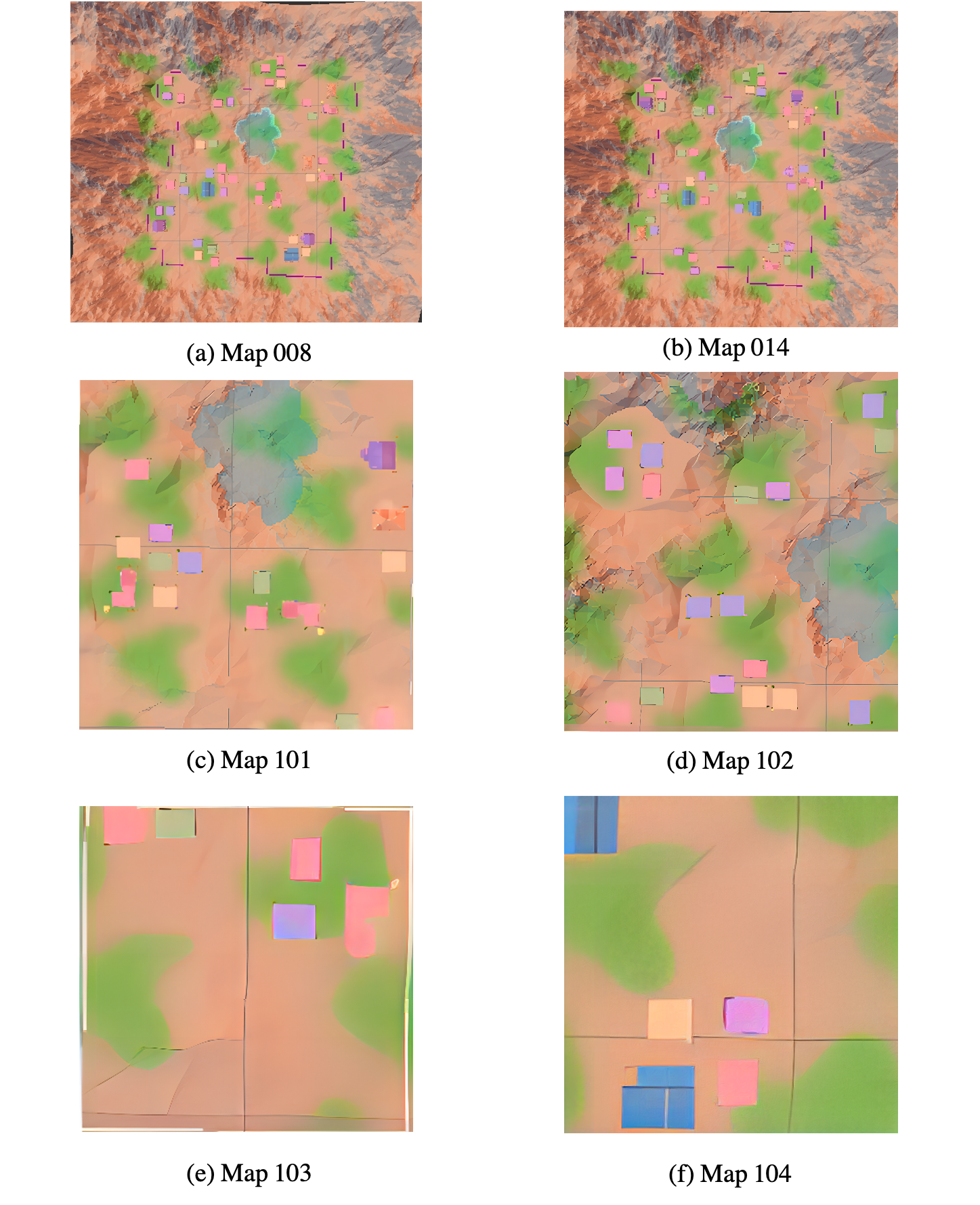}
    \caption{Different maps visualization. We can see different terrain shapes and different housing density. For map 008 and map 014, we select two big maps. For map 101,102,103,104, we randomly generated a big map then cut out a small area of the big map. }
    \label{mapviz}
\end{figure}

To demonstrate the potential in using our benchmark environment to learn robust agent policies with reasonably good capacity of generalization, we created the training maps with different sizes, house numbers, house densities and storey numbers. For each size of the map, we first select the number of houses, the house density and the range of the number of floors of a house. Then we randomly sample from all the legal locations where to generate the houses based on the selected density. After this, specific building structure of the house and the storey numbers are randomly selected from the predetermined asset pool. In this way, we can easily generate thousands of different maps. But in this paper we focus on a small scale evaluation where we only generate 6 maps of 3 sizes to compare the performance of different algorithms on various tasks. Figure~\ref{mapviz} shows the top view of the generated maps and the detailed generation configurations of them are given in Table~\ref{mapconfig}.

\begin{table}[htbp]
\label{mapconfig}
	\centering
	\caption{Generation configurations of the benchmark maps}\label{winningrate}
	\begin{tabular}{cccccc}
	    \toprule
		Map ID & Size & House numbers & House density & Storey numbers  \\
		\midrule
		008 &  500 $\times$ 500 &12&low&2-3\\ 
		014 &  500 $\times$ 500 &15&high&2-3\\ 
		101 & 200 $\times$ 200 &4 &low&1 \\ 
		102  &  200 $\times$ 200 &8& high&2-3\\ 
		103 &  100 $\times$ 100 &2&low&1\\ 
		104 &  100 $\times$ 100 &4&high&2-3\\ 
		\bottomrule
	\end{tabular}
	\label{neighbor}

\end{table}

\subsection{Computation cost and simulation speed}

We also evaluate the simulation speed of different number of workers and agents, this result is shown in table \ref{simspeed}. All the results are tested on a single AMD 5800H CPU core.



\begin{table}[!htbp]
\label{simspeed}
	\centering
	\caption{Throughputs (total FPS) of the environment when running on a single CPU machine with different numbers of concurrent process and agents. The maximum throughput for each number of agents is bolded.}\label{winningrate}
	\begin{tabular}{|c|c|c|c|c|c|c|c|}
        \hline
        \multicolumn{2}{|c|}{\multirow{2}*{throughput}} & \multicolumn{6}{c|}{\# process} \\
        \cline{3-8}
        \multicolumn{2}{|c|}{} & 1 & 2 & 4 & 6 & 8 & 10 \\
        \hline
        \multirow{3}*{\# agents} & 1 & 499.61 & 996.90 & 1974.84 & 2766.21 & 3506.31 & \textbf{4187.03} \\
        \cline{2-8}
        & 5 & 500.18 & 996.94 & 1609.81 & 1463.60 & 1864.10 & \textbf{2023.76} \\
        \cline{2-8}
        &10&500.54&997.62&1095.73&960.10&\textbf{1426.10}&1246.24 \\
        \hline
	\end{tabular}
	\label{neighbor}

\end{table}

\subsection{Reward Configuration}
We designed the reward functions for the three evaluation tasks simply by translating the agent state into a binary reward of whether reaching the goal. The detailed operations to compute the reward are given below in a form of Python code snippets.

\begin{itemize}
\item Single-Agent Navigation
\begin{python}
state = self.update_state()
curr_loc = self.get_location(state)
targ_loc = self.target_location
if self.get_distance(curr_loc, targ_loc) <= 1:
    reward = 1
else:
    reward = 0
\end{python}

\item Cooperative task 1: supply gathering
\begin{python}
state = self.get_states(agent_id)
if state.num_supply > self.collected_supply[agent_id]:
    rewards[agent_id] = 1
    self.collected_supply[agent_id] = state.num_supply
else:
    rewards[agent_id] = 0
\end{python}




\item Cooperative task 2: target capture
\begin{python}
state = self.get_states(agent_id)
curr_loc = self.get_location(state)
targ_loc = self.target_location
if self.get_distance(curr_loc, targ_loc) <= 1:
    rewards[agent_id] = 1
else:
    rewards[agent_id] = 0
\end{python}



\end{itemize}

\subsection{Environment interface}
WILD-SCAV can support popular Gym interface~\cite{brockman2016openai}. The typical interface is shown below. To facilitate the design of new algorithm and benchmark comparison, our gaming interface can support different maps, different locations, different reward design for different tasks.  

\begin{python}
game = Game(
    map_dir="/path/to/map_data",
    engine_dir="/path/to/engine_backend",
)
game.init()

env = gym.make(
    id="WildScav-Navigation-v0",
    env_config={
        "engine": game 
        "map_id": 1,
        "timeout": 30,
        "start_location": [0, 1, 0],
        "target_location": [5, 0, 3],
        obs=[depth_image, position],
        reward= RewardManager.add('Navigation')
    }
)
\end{python}

\subsection{ Model description}

We also provide the training details of the RL algorithms on WILD-SCAV using standard RLlib library~\cite{liang2018rllib}. We run all experiments on a Linux server with 96 CPU cores. We fix the number of experience sampler workers to $80$ (1 CPU per worker) and allocate the rest of the CPUs to the master learner. We use Adam as the default optimizer for all models trained in the experiments. The detailed training configurations for experiments on PPO, A3C, IMPALA can be found in Table ~\ref{ppoconfig}, Table~\ref{a3cconfig}, Table~\ref{impalaconfig} respectively.

\begin{table}[htbp]
\centering
\caption{Training configuration of PPO experiments}
\begin{tabular}{ccc}
    \toprule
    Parameter Name  &  Value   \\
    \midrule
    learning rate & 0.00001 \\ 
    train batch size &  32000 \\ 
    SGD mini-batch size & 256 \\ 
    rollout fragment length &  400 \\ 
    entropy coefficient & 0\\ 
    value loss coefficient & 1.0 \\
    KL-divergence coefficient & 0.2 \\
    \bottomrule
\end{tabular}
\label{ppoconfig}
\end{table}

\vspace{5mm}

\begin{minipage}[htbp]{\textwidth}

\begin{minipage}[t]{0.48\textwidth}
\makeatletter\def\@captype{table}
\centering
\begin{tabular}{ccc}
\toprule
Parameter Name  & Value   \\
\midrule
learning rate  &  0.0001 \\ 
rollout fragment length  &  400 \\ 
entropy coefficient &  0.01 \\ 
value loss coefficient & 0.5 \\
\bottomrule
\end{tabular}
\label{a3cconfig}
\caption{Training configuration of A3C}
\end{minipage}
\begin{minipage}[t]{0.48\textwidth}
\makeatletter\def\@captype{table}
\centering
\begin{tabular}{ccc}
\toprule
Parameter Name  & Value   \\
\midrule
learning rate & 0.0005 \\ 
train batch size & 32000 \\
rollout fragment length & 400 \\
entropy coefficient & 0.01 \\
value loss coefficient & 0.5 \\
\bottomrule
\end{tabular}
\label{impalaconfig}
\caption{Training configuration of IMPALA}
\end{minipage}
\end{minipage}

\subsection{ Network architecture description}

We use the following network architecture configuration to encode the information. 

\begin{python}[htbp]
    # FullyConnectedNetwork
    # Number of hidden layers to be used.
    "fcnet_hiddens": [256, 256],
    # Activation function descriptor.
    # Supported values are: "tanh", "relu", "swish" (or "silu"),
    # "linear" (or None).
    "fcnet_activation": "tanh",

    # VisionNetwork
    # If None, automatically determines the CNN filter shapes based on the input size
    "conv_filters": None,
    # Activation function descriptor.
    # Supported values are: "tanh", "relu", "swish" (or "silu"),
    # "linear" (or None).
    "conv_activation": "relu",
\end{python}

\subsection{Running time}

\begin{table*}[htbp]
    \centering
    
    \begin{tabular}{c|c|c|c}
    
    \toprule
    \multicolumn{2}{c}{\textbf{A3C}} & \multicolumn{1}{c}{\textbf{PPO}} & \multicolumn{1}{c}{\textbf{IMPALA}} \\
    
    \midrule
    Map  & Running time (s)          &   Running time  (s)      & Running time(s)   \\
    \midrule
    101  & 10456.62   & 29414.05 & 4470.43 \\
    102  & 10760.02   & 28100.36 & 4255.84 \\
    103  & 16922.06  & 31898.47 & 6455.39 \\
    104  & 1962.49  & 9855.69 & 7014.64 \\
    008  & 6076.44   & 28854.93 & 4211.52 \\
    014  & 6352.47   & 28305.63 & 4278.53 \\
	\bottomrule
    
    \end{tabular}
    
    \caption{Evaluation results on the navigation task. We report  the total running time in second.}\label{coop2table}
    
    \end{table*}

\begin{table*}[htbp]
    \centering
    
    \begin{tabular}{c|c|c|c}
    
    \toprule
    \multicolumn{2}{c}{\textbf{A3C}} & \multicolumn{1}{c}{\textbf{PPO}} & \multicolumn{1}{c}{\textbf{IMPALA}} \\
    
    \midrule
    Map  & Running time (s)          &   Running time  (s)      & Running time (s)  \\
    \midrule
    101  & 1925.99   & 12250.53 & 2312.53\\

	\bottomrule
    
    \end{tabular}
    
    \caption{Evaluation results on the cooperative supply gathering. We report  the total running time in second.}\label{coop2table}
    
    \end{table*}

\begin{table*}[htbp]
    \centering
    
    \begin{tabular}{c|c|c|c}
    
    \toprule
    \multicolumn{2}{c}{\textbf{A3C}} & \multicolumn{1}{c}{\textbf{PPO}} & \multicolumn{1}{c}{\textbf{IMPALA}} \\
    
    \midrule
    Map  & Running time (s)         &   Running time   (s)     & Running time (s)  \\
    \midrule
    101  & 1300.56   & 16885.55 & 1748.60\\

	\bottomrule
    
    \end{tabular}
    
    \caption{Evaluation results on the cooperative target capture. We report  the total running time in second.}\label{coop2table}
    
    \end{table*}

\section{Winning Policy Analysis}

We held an online competition in the past few months to test the RL algorithm for solving this problem. 
Based on our previous results in the competition~\footnote{\url{https://sites.google.com/view/inspirai-wildscav-cog2022/home}}, we analyze the winning policy in the leader board below. Compared to the rule-based method, such as A star algorithm~\cite{hu2012pathfinding}. We can find that the traditional reinforcement learning agent cannot efficiently handle this task. Therefore, we analyze the winning strategies from the top submissions. 

\subsection{Navigation}

The goal of this task is to make the agent reach the target as soon as possible. The challenge is that in this track, the spawn point and destination location basically do not involve indoors, but the agent still has the possibility of entering the room by mistake. Some destinations are set on hillsides, and the agent needs to know how to climb hills. 

After testing the map, we found that the extreme value of the boundary coordinates does not exceed ±300, and the tallest building does not exceed four floors. Therefore, we established a 600*600*4 obstacle marker matrix for sampling and recording terrain information. There are several tricks to solve these tasks:

\begin{itemize}
    \item During the movement of the agent, the current position will be stored as a checkpoint every time given a certain number of time steps. If the agent fails to move a valid distance for a long time, it will try to return to the previous saved checkpoint.
    
    \item When we use visual depth image to detect an obstacle, the slope of the obstacle is also calculated. We assume there is no obstacle if the slope is low.
    
    \item When the agent is walking, if there is obstacle but there is no obstacle in the image, we will try to jump.
    
\end{itemize}

\subsection{Supply gathering}

In the second task, the agent needs to collect as many supplies as possible on the map in a fixed time. There are a large number of indoor scenes on the map, which requires the agent to have the ability to explore complex indoor scenes.
The buildings on the map have different floors, and the spawn point may also be located on the upper floor. There are several tricks to solve this task:
\begin{itemize}
    \item The supply gathering problem can be decomposed into multiple navigation sub-problems. In each step, we maintain a list of information, such as supply points, and location information, to make the decision.
    \item The agent will use a decision tree policy to determine when and where to collect supply. 
    \item In the high-floor area, the agent will record the exploration route and return to the original route first when it is necessary to go downstairs.
\end{itemize}

\subsection{Supply battle}

In the third task, compared with supply gathering, multiple agents are competing on the same map, and it is necessary to consider killing other agents to gain more resources. There are a large number of indoor scenes on the map, which requires the agent to have the ability to explore complex indoor scenes. The buildings on the map have different floors, and the spawn point may also be located on the upper floor. Agents need to have certain attacking capabilities, such as shooting other agents.

\begin{itemize}
    \item If the agent detects local information, it will automatically aim at the nearest enemy and fire.
    \item We will consider reloading when the current clip is lower than a threshold.

\end{itemize}

\section{Further information about our gaming enginee}

Our environment consists of two major components, the game engine (backend) and gameplay API (frontend). We have described the gameplay interface and training algorithm configurations in previous sections. In this section, we provide more information on how we build the infrastructure and optimize the simulation throughput. 

\textbf{Communication.} we use GRPC \footnote{\url{https://grpc.io/}} protocol to design the format and content of observation, action, environmental control command and event messages. The protocol will convert data from both ends into messages with a uniform format so that they can be transmitted between processes across multiple languages. To ensure synchronous control of the game simulation, we built a simple message queue server on the frontend side to receive the observation message from the engine, wait for the algorithm level to consume, and then send back the new action to the engine.

\textbf{Simulation optimization.} We consider the following strategies to optimize the gaming performance:
\begin{itemize}
    \item The game scene is too large, and there are too many resources, resulting in excessive rendering pressure and causing stuttering. As a result, we reduce the rendering batch of the game from more than 3000 to an average of around 200.
    
    \item On the premise of meeting the requirements of the algorithm, the communication request of GRPC is reduced. Therefore, the running speed of the engine is improved, and the training rate is accelerated.
    
    \item We optimize the management of the protocol data to reduce running consumption caused by game memory management.
    
\end{itemize}

\end{document}